\documentclass[letterpaper, 10 pt, conference]{ieeeconf}  

\def\baselinestretch{0.8503}\selectfont
\usepackage{cite}
\usepackage{amsmath,amssymb,amsfonts}
\usepackage{algorithmic}
\usepackage{graphicx}
\usepackage{textcomp}
\usepackage{xcolor}
\usepackage{makecell}
\usepackage{float}
\usepackage{subcaption}
\usepackage{caption}
\usepackage{titlesec}

\titlespacing*{\section}
{0pt}{3pt}{3pt}
\titlespacing*{\subsection}
{0pt}{2pt}{2pt}
\titlespacing*{\subsubsection}
{0pt}{1pt}{1pt}
\usepackage{booktabs}

\def\BibTeX{{\rm B\kern-.05em{\sc i\kern-.025em b}\kern-.08em
    T\kern-.1667em\lower.7ex\hbox{E}\kern-.125emX}}
\IEEEoverridecommandlockouts                              

\title{\LARGE \bf
Semantic Feature Matching for Robust Mapping in Agriculture
}

\author{Mohamad Qadri and George Kantor
\thanks{This work was supported in part by the DOE APRAE TERRA 3P project, and USDA NIFA CPS 2020-67021-31531}
\thanks{Mohamad Qadri and George Kantor are with The Robotics Institute, Carnegie Mellon University, 5000 Forbes Avenue, Pittsburgh PA, USA. {\tt\small mqadri@andrew.cmu.edu, kantor@ri.cmu.edu}}
}
\begin{document}

\maketitle
\thispagestyle{empty}
\pagestyle{empty}

\begin{abstract}
Visual Simultaneous Localization and Mapping (SLAM) systems are an essential component in agricultural robotics that enable autonomous navigation and the construction of accurate 3D maps of agricultural fields. However, lack of texture, varying illumination conditions, and lack of structure in the environment pose a challenge for Visual-SLAM systems that rely on traditional feature extraction and matching algorithms such as ORB or SIFT. This paper proposes 1) an object-level feature association algorithm that enables the creation of 3D reconstructions robustly by taking advantage of the structure in robotic navigation in agricultural fields, and 2) An object-level SLAM system that utilizes recent advances in deep learning-based object detection and segmentation algorithms to detect and segment semantic objects in the environment used as landmarks for SLAM. We test our SLAM system on a stereo image dataset of a sorghum field. We show that our object-based feature association algorithm enables us to map 78\% of a sorghum range on average. In contrast, with traditional visual features, we achieve an average mapped distance of 38\%. We also compare our system against ORB-SLAM2, a state-of-the-art visual SLAM algorithm.
\end{abstract}


\section{Introduction}
Simultaneous Localization and Mapping (SLAM) enables autonomous robot navigation in unknown environments by providing a joint estimate of the robot poses and the 3D location of landmarks. SLAM has important applications in agriculture since the 3D reconstructions obtained from mapping agricultural fields can provide valuable information for downstream tasks such as plant phenotyping, crop counting, and yield prediction. One such application is high-throughput sorghum seed phenotyping. Sorghum is an essential grain for food and energy production, and significant efforts are being put into utilizing phenotyping to improve crop quality and understand sorghum genetic variations \cite{genome}.\par
Stereo-based visual systems are often the preferred choice when imaging agricultural fields because of their wide availability and reasonable cost. Stereo cameras can also resolve finer details compared to LIDAR, which only provides a sparse representation of the environment \cite{rols}.
However, existing visual-based SLAM and 3D reconstruction algorithms such as ORB-SLAM2 and Structure from Motion (SFM) pipelines that fundamentally rely on the accuracy of traditional visual feature matching algorithms such as SIFT and ORB  often fail. These failures are due to the lack of texture in the images, variations in luminosity levels, the dynamics of the environment (for example, leaves or crops moving due to the wind), and the presence of repeated patterns. \par
\begin{figure}[h!]
\centerline{\includegraphics[width=1\linewidth]{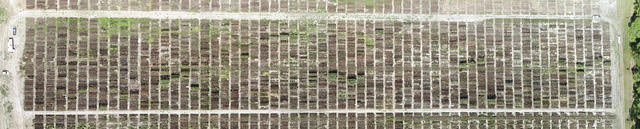}}
\caption[]{\textit{Aerial views of a sorghum breeding panel at Clemson University's Pee Dee Research and Educational Center in Florence, SC. The field is organized into plots, where each plot is approximately $2m \times 5m$ and contains a different variety of sorghum}}
\label{aerial2}
\end{figure}

On the other hand, Robotic navigation in agricultural fields exhibit a specific structure: farms are composed of rows arranged in a straight line (Fig. \ref{aerial2}), and robots traverse the field one row at a time, ideally moving in a straight line while taking images of plants from the sides
\cite{robotanist}. 
This paper aims at taking advantage of this structure and presents a SLAM system and an object-level data association algorithm that use the imaged sorghum seeds as semantic  landmarks for SLAM. Our data association algorithm considers the detected seeds in an image as nodes in a graph and makes the following observations: 1) the position of each node relative to its surrounding nodes should stay approximately constant across images. 2) the difference in 2D coordinates of the same node appearing in two consecutive images (or in a stereo image pair) is predominantly defined by a horizontal translation (Fig. \ref{fig:series}). Our SLAM system was tested on a stereo-image dataset of a sorghum field and can easily be extended to other types of crops. 
\newline
\textbf{The contributions of this paper are twofold:}
\begin{itemize}
    \item  We propose an object-level feature association algorithm that enables the creation of 3D reconstructions robustly by taking advantage of the structure of robotic navigation in agricultural fields.
    \item An object-level SLAM system that utilizes recent advances in deep learning-based object detection and segmentation algorithms to detect and segment semantic objects in the environment which we use as landmarks for SLAM. Our SLAM system does not use inertial sensory measurements and only relies on visual odometry from a stereo camera capturing images at the low frame rate of 5Hz.
\end{itemize}
\section{Related Work}
\subsection{SLAM}
Current state of the art indirect feature-based Visual-SLAM algorithms, such as ORB-SLAM2 \cite{orbslam2}, depend on the accuracy of the extracted features, such as ORB, and can fail in environments where feature matching is not robust. Direct SLAM methods such as LSD-SLAM \cite{lsd-slam} operate directly on pixel-level brightness information and achieve tracking and pose estimation by minimizing the photometric error between the new frame and the current keyframe. However, direct methods are more susceptible to changes in lighting conditions,  common in images taken in an agricultural field. Several works exist on object-level SLAM: SLAM++ \cite{slam++} is an RGB-D based system that detects known objects in the environment, which are then used as high-level landmarks. However, SLAM++ requires the creation of a database of high-quality 3D models used as prior. The authors in \cite{choudhary2014slam} proposed a method for the online discovery of objects in indoor environments without the use of prior models. Each object is represented by the centroid of its associated pointcloud. However, only non-planar regions are considered as potential landmarks which limits the method's applicability in our setting: From the viewpoint of our camera, sorghum panicles appear to be mostly planar. In addition, our objective is to use individual seeds as landmarks and not the entire sorghum panicle. With the progress made in 2D object detection and segmentation, numerous works combine deep learning and SLAM: Fusion++ \cite{fusion++} utilizes Mask-RCNN \cite{Mask-rcnn}  to initialize Truncated
Signed Distance Function (TSDF) reconstructions, which are, in turn, used as object-level representations of landmarks.  Fusion++ mostly targets indoor environments.
QuadricSLAM \cite{quadric} and CubeSLAM \cite{cubic} are monocular-based SLAM systems. They propose two different 3D landmarks representations, ``quadrics" and ``cuboids," respectively, which are estimated from 2D bounding boxes without prior models. They are based on a factor-graph based SLAM formulation that jointly estimates the 3D pose of these objects along with the camera poses. The authors in \cite{rols} explain that these approaches' performance will be poor in cluttered environments such as a sorghum field where there is a significant overlap between bounding boxes. Similarly, ROSHAN \cite{ok2019robust} introduced an ellipsoid landmark-based monocular SLAM system which constraints the ellipsoid representation using information from detected bounding boxes, a prior on the shape of the object, and a ``texture" plane estimated from triangulated feature points on the surface of the object. In our application, the overlap of the detected seeds' bounding boxes and the lack of texture in the images limit the applicability of such methods. Kimera \cite{kimera} uses Visual-Inertial Odometry to estimate the state of the robot and builds a semantic mesh model of the environment. We note that our system focuses on Stereo-Visual Odometry. \par
Most state-of-the-art SLAM algorithms are framed as factor graph optimization problems where nodes represent robot poses or 3D landmarks in the environment, and edges (factors) are probabilistic constraints on these variables derived from sensor measurements. We use GTSAM \cite{gtsam} as the underlying optimization framework, a library optimized for solving sparse factor graph representations. We refer the reader to \cite{explained} for a detailed treatment of factor graphs. 
\vspace{2px}
\subsection{2D Object Detection and Segmentation} 
Several impressive 2D object detection (YOLOv3 \cite{yolov3}, SSD \cite{ssd}, Faster-RCNN \cite{Faster-RCNN}) and semantic segmentation networks (PSPNet \cite{pspnet}, Mask-RCNN \cite{Mask-rcnn}) based on deep neural networks, have recently been proposed. StalkNet \cite{stalknet} proposed an image processing pipeline used to measure stalk width. It used a Faster-RCNN network for image detection and a Fully Convolutional Network (FCN) to segment stalks. Ellipses are then fitted to the segmented stalks and used to measure their width. \cite{parharcgan} improved the segmentation accuracy of StalkNet by using an image-to-image translation conditional adversarial network ``pix2pix" \cite{Pix2Pix} as the segmentation network. 

\section{Proposed Method}
\subsection{System Overview}
We describe our system's main blocks shown in Fig. \ref{fig:pipeline}: the Detection-and-Segmentation block takes an image as an input and outputs a set of extracted semantic keypoints. The Object-Level-Feature-Association block performs temporal or stereo matching between two sets of semantic keypoints. Iterative Closest Point (ICP) is used to determine the relative transformation between two frames and hence, an estimate of the current camera pose. Initial estimates of the 3D landmarks and camera poses are used as the initial guess for a factor graph-based non-linear optimization, which returns estimates of the camera poses and landmarks' 3D location up to time $t$.
\subsection{Feature Extraction}
\begin{figure}[H]
\centerline{\includegraphics[width=1\linewidth, height=1.35in]{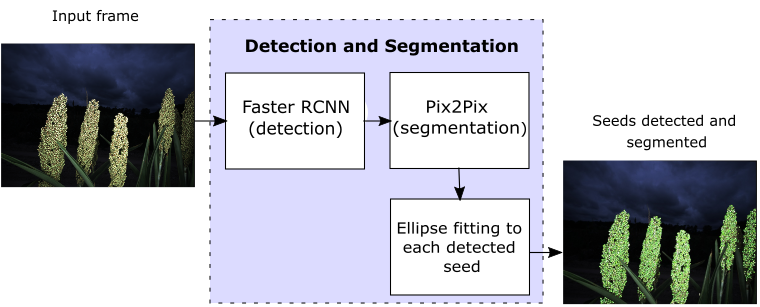}}
\caption{\textit{Our detection and segmentation pipeline. The center of the each ellipse is considered the projection of a 3D landmark.}}
\label{detect_seg}
\end{figure}

Our feature extraction pipeline (see Fig. \ref{detect_seg}) is based on \cite{parharcgan}. A Faster-RCNN network with a VGG16 backbone is used as the detection network and returns a bounding box for each seed seen in the image. Each bounding box is cropped out and passed to the pix2pix network, which generates a new image with a segmentation mask for the detected seed. An ellipse is then fitted to the segmented area, and the ellipse's center is used as one semantic keypoint. We trained the Faster-RCNN and pix2pix on an Nvidia RTX 2080 GPU.
\vspace{-2.5mm}
\subsection{Data association algorithm}
We run the detection and segmentation pipeline at time $t$ on both the left and right stereo images. We obtain two sets of points: $C^{L}_{t}$, which is the set of 2D coordinates corresponding to the centers of the sorghum seeds that appear in the left image, and similarly, $C^{R}_{t}$ which we obtain from the right image. In stereo matching, the objective is to associate each point in $C^{L}_{t}$ to its more likely image in $C^{R}_{t}$. In temporal matching, the aim is to associate each point in $C^{L}_{t-1}$ to its more likely image in $C^{L}_{t}$. Our data association algorithm is agnostic to what type of matching we are performing since the objective is the same in both cases: assign each 2D point in a set U to its more likely pair in another set V.
\begin{figure}[h!]
\centerline{\includegraphics[width=1\linewidth]{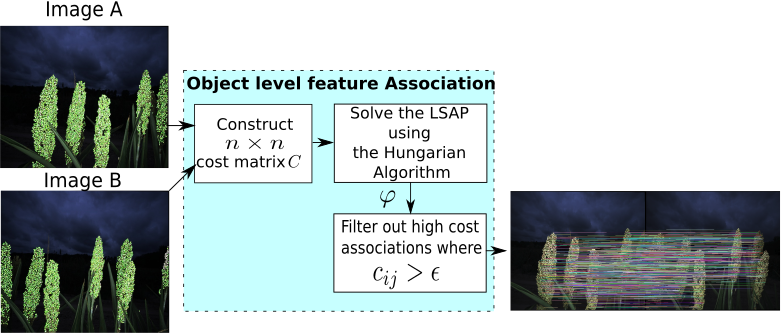}}
\caption{\textit{Proposed feature association pipeline.}}
\label{fig:data_association}
\end{figure}
\begin{figure*}[h!]
\begin{center}
   \includegraphics[width=0.97\linewidth, height=3in]{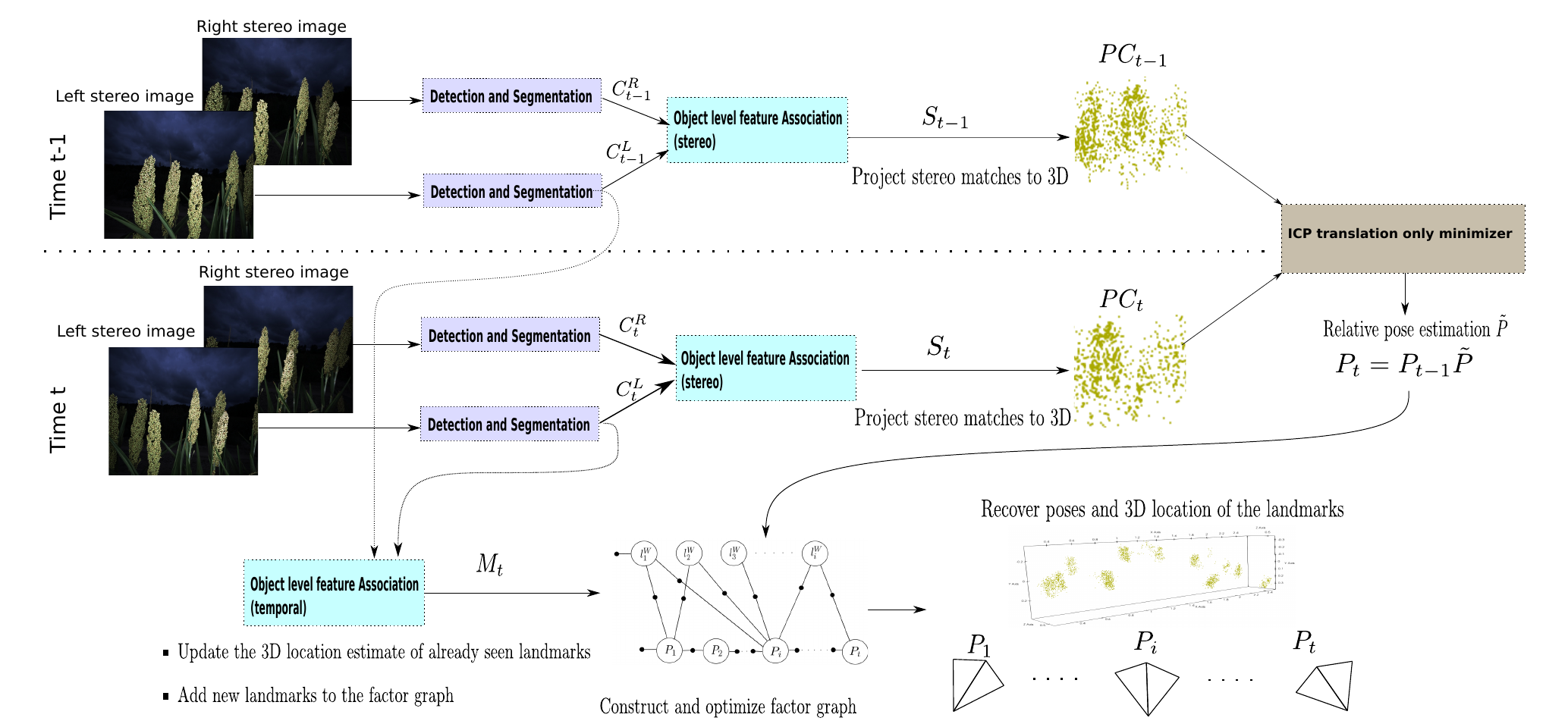}
\end{center}
   \caption{\textit{System Overview}}
\label{fig:pipeline}
\end{figure*}
\subsubsection{Defining the Linear Sum Assignment Problem}
The object-level data association algorithm between stereo pairs or successive temporal frames is framed as a \textit{linear sum assignment problem} (LSAP) optimization problem \cite{assignment}. 
We define a bipartite graph $G = (U, V, E)$. Each vertex $s_{ab} \in U$, with coordinate $(a,b)$ in 2D camera frame, corresponds to the projection of a 3D landmark onto image A. Similarly, each vertex $s_{xy}$ with coordinates $(x,y)$ $ \in V$, corresponds to the projection of a landmark  onto image B. An edge $c_{ij} \in E$ between nodes $s_{ab}$ and $s_{xy}$ defines the cost of associating $s_{ab}$ to $s_{xy}$.
By introduction an assignment matrix $\varphi$, LSAP can be framed as the following optimization problem:
\begin{align*}
    \min_{\varphi \in S_{n}} \sum_{i=1}^{n} \sum_{j=1}^{n} c_{ij} \varphi_{ij} \
        \text{subject to}:    \sum_{j=1}^{n} \varphi_{ij} = 1 ,\ i \in U \\
     \sum_{i=1}^{n} \varphi_{ij} = 1 ,\ j \in V \\
     \varphi_{ij} \in \{0,1\}
\end{align*}
Where $S_{n}$ is the set of all possible assignment of nodes in  $U$ to nodes in $V$.
\vspace{2px}
\subsubsection{Cost function}
\vspace{2px}
Since sorghum panicles are rigid bodies in space, the distance from a particular seed to its neighboring seeds should stay approximately constant as the robot moves in the environment. In terms of our graphical model, we add the constraint that the sum of the Euclidean distances of a node $s_{ab} \in U$ to its surrounding nodes in $U$ should be approximately equal to the sum of the Euclidean distances of the [image of $s_{ab}] \in V$ and its surrounding nodes in $V$. We define a heuristic cost function that captures this relative geometric structure between the landmarks: 
\newline
For each node $s_{ab} \in U $, define four sets of neighbouring nodes: $L_{ab}$ (left), $R_{ab}$ (right), $T_{ab}$ (top), and $B_{ab}$ (bottom) satisfying the following conditions:
\begin{align*}
L_{ab} = \{\forall s'=(c,d) \in U \: | \: 0 < a - c < \Delta \: \text{and} \: |d-b| < \epsilon\} \\ 
R_{ab} = \{\forall s'=(c,d) \in U \: | \: 0 < c - a < \Delta \: \text{and} \: |d-b| < \epsilon\} \\ 
T_{ab} = \{\forall s'=(c,d) \in U \: | \: 0 < b-d < \Delta \: \text{and} \: |c-a| < \epsilon\} \\ 
B_{ab} = \{\forall s'=(c,d) \in U \: | \: 0 < d-b < \Delta \: \text{and} \: |c-a| < \epsilon\}
\end{align*}
We define $L_{xy}$, $R_{xy}$, $T_{xy}$ and $B_{xy}$ similarly for $s_{xy} \in V $. The cost of associating node $s_{ab}$ to node $s_{xy}$ is defined as:
\begin{align}
    C(s_{ab}, s_{xy})  = r \cdot C'(L_{ab}, L_{xy}) +  r \cdot C'(R_{ab}, R_{xy}) +  \nonumber \\
    r \cdot C'(B_{ab}, B_{xy})+ r \cdot C'(T_{ab},T_{xy}) + |b-y| 
\end{align}

\begin{align*}
   \text{where: }C'(X, Y) = \frac{\sum_{s' \in X} \sqrt{(s'(x) - a)^2 + (s'(y) - b)^2}}{\sum_{s' \in Y} \sqrt{(s'(x) - x)^2 + (s'(y) - y)^2}}, \\ 
\end{align*}
$r$ is a constant, and $|b - y|$ is a term added to penalize matching landmarks with high vertical difference $|b-y|$. $s'(x)$ and $s'(y)$ are respectively the $x$ and $y$ coordinate of one of the surrounding nodes. The Hungarian algorithm \cite{assignment} solves the LSAP in $O(n^3)$. Efficient CUDA-based implementations running on Nvidia GPUs  \cite{gpuhungarian2} provide significant speedups compared to CPU-based implementations.
\vspace{2px}
\subsubsection{Cost as a matching confidence measure}
\vspace{2px}
The Hungarian algorithm returns an assignment matrix $\varphi$, which is a bijection from $U$ to $V.$ Row $i$ in $\varphi$ contains all zeros except at a column $j$ where $\varphi_{ij}=1$ meaning that node $i$ has been matched with node $j$. This assignment corresponds to the cost $c_{ij}$. Filtering out assignments where $c_{ij} > \epsilon$ allows us to keep only high confidence matches (matches with low cost).
\vspace{1px}
\subsection{3D pose estimation}
Given stereo matches $S_{t} = \{x_{t}, y_{t}, u_{t}\}$ at time $t$ and stereo matches $S_{t-1} = \{x_{t-1}, y_{t-1}, u_{t-1}\}$ at time $t-1$ where $u$ refers to the $x$ coordinate of the matched keypoint in the right stereo image, two pointclouds $PC_{t-1} = \{ \mathbf{p_{i}} \}$ and $PC_{t} = \{ \mathbf{q_{j}} \}$, where $\mathbf{p_{i}}, \ \mathbf{q_{j}} \in \mathbb{R}^3$,  can be obtained by unprojecting $S_{t}$ and $S_{t-1}$ respectively to 3D.
\begin{align}
    p^{x}_{i} = \frac{(x_t - c_{x}) \cdot p^{z}_{i}}{f}; \,
    p^{y}_{i} = \frac{(y_t - c_{y}) \cdot p^{z}_{i}}{f}; \, 
    p^{z}_{i} = \frac{b \cdot f}{d}
\label{eq:proj}
\end{align}
and similarly for $\mathbf{q_j}$. $b$ refers to the stereo baseline, $d = x - u_{R}$ is the disparity in pixels and, $f$, $c_{x}$, and $c_{y}$ refer to the intrinsic camera parameters. The relative pose transformation $\tilde{P} \in SE(3)$ between the previous and current frame is obtained by minimizing the translation between the two pointclouds and assuming zero rotation (since the robot moves in a straight line while capturing stereo images from the sides, the transformation should be defined mostly by a translation). The translation vector is then projected onto the direction of motion (known as a prior). That is: 
\begin{align}
  &\tilde{P} = \begin{bmatrix} \textbf{I}_{3\times3} & \Pi{(\tilde{\textbf{t} })}\\ 
   \textbf{0} & 1 \\
   \end{bmatrix} \text{with } \Pi \text{ being the orthogonal projection} \nonumber  \\
   & \text{operator, and  }\tilde{\textbf{t}} \text{ found by minimizing the error term:}   \nonumber \\
   & E =  \sum_{i=1}^{N}  ||\tilde{\textbf{R}} \textbf{p}_{i} + \textbf{t} - \textbf{q}_{i} ||^2 \quad \text{with} \quad \tilde{\textbf{R}} = \textbf{I}_{3\times3} 
\end{align}
 The estimated camera pose at time t is then $ P_{t} = P_{t-1} \tilde{P}$.
This variant of point-to-point ICP has been implemented in Libpointmatcher \cite{libpointmatcher}.
\subsection{Factor Graph Optimization}
\begin{figure}[H]
\centerline{\includegraphics[width=0.7\linewidth]{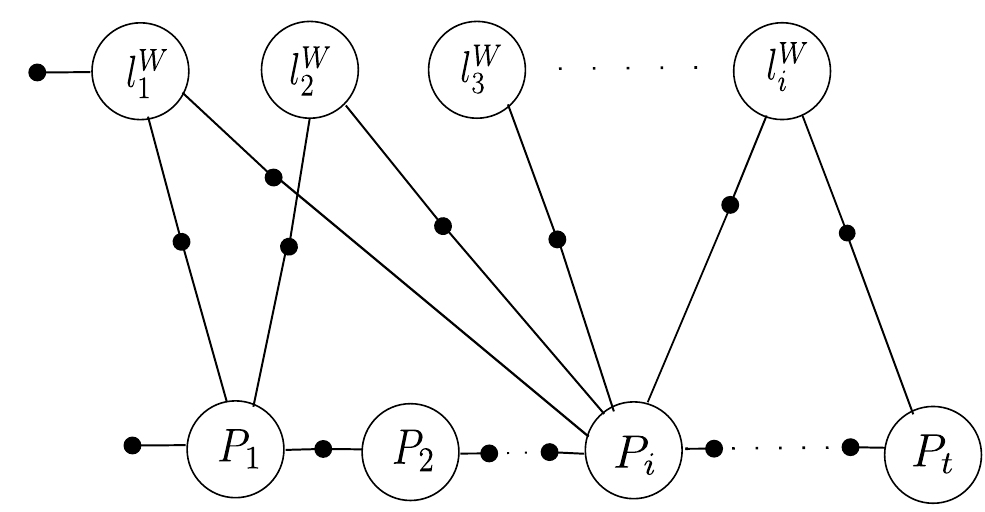}}
\caption{\textit{Factor graph representation}}
\label{fig:factor_graph}
\end{figure}
We use our data association algorithm to find temporal matches $M_{t}$ between the previous and current frame: for every landmark seen in the left stereo image at time $t-1$, we find its corresponding 2D projection in frame $t$.  
\newline 
We add a landmark to the factor graph if it is observed by at least two different camera poses. To obtain a good initial guess of the 3D location of landmark $i$ observed at time $t$, we average out all its previous 3D location estimates as follows:
\begin{itemize}
    \item If landmark $i$ has been observed from exactly one previous camera pose $P_j$ where $ j \in [1 \hdots t-1]$, 
    we set its 3D coordinate estimate in world reference frame to
    \begin{equation}
        l^W_{i} = \frac{PC^W_{j}[i] + PC^W_{t}[i]}{2}  \in \mathbb{R}^3
    \end{equation}
    where $PC^W_{j}[i]$ and $PC^W_{t}[i]$ are obtained by transforming the landmark's 3D location estimates $PC_{j}[i]$ and $PC_{t}[i]$ from their respective camera frames to the world reference frame.
    \item If landmark $i$ has been seen from $N > 1$ previous poses in $[P_{1} \hdots P_{t-1}]$, we extract out its 3D optimized coordinate ($l^{W}_{optimized}$), from the previous optimization step and set the new 3D coordinate estimate of landmark $i$ to:
    \begin{equation}
         l_{i}^{W} = \frac{N \cdot l^{W}_{optimized} +  PC^W_{t}[i]}{N+1} \in \mathbb{R}^3
    \end{equation}
\end{itemize}
   
The estimates of the 3D landmark location and camera pose are used as the initial guess for the non-linear optimization. We construct the non-linear factor graph in GTSAM at time $t$ as shown in Fig. \ref{fig:factor_graph}. Each factor in the factor graph is a stereo projection factor, a constraint between pose $P_{i}$ and a landmark $l^W_{j}$. It is composed of a $\textit{StereoPoint2}(x, u_{R}, y)$ where $(x,y)$ is the projection of landmark $l^W_{j}$ onto the image plane at pose $P_{i}$, and $u_{R}$ is the $x$ coordinate of the landmark in the corresponding right stereo image taken at pose $P_{i}$. We use the Huber robust noise model which allows for modeling outliers. The estimated pose $\tilde{P}$ between consecutive frames also specifies a motion constraint between the two camera poses $P_{i}$ and $P_{i+1}$. The motion noise model is tuned according to our prior on camera motion (i.e we specify a small uncertainty in the rotation and high uncertainty in the direction of motion $\Pi(\tilde{t})$). The factor graph is optimized using a Dogleg batch optimizer.
\vspace{1px}
\subsection{Pointcloud Postprocessing}
\begin{figure}[h!]
\centerline{\includegraphics[width=1\linewidth]{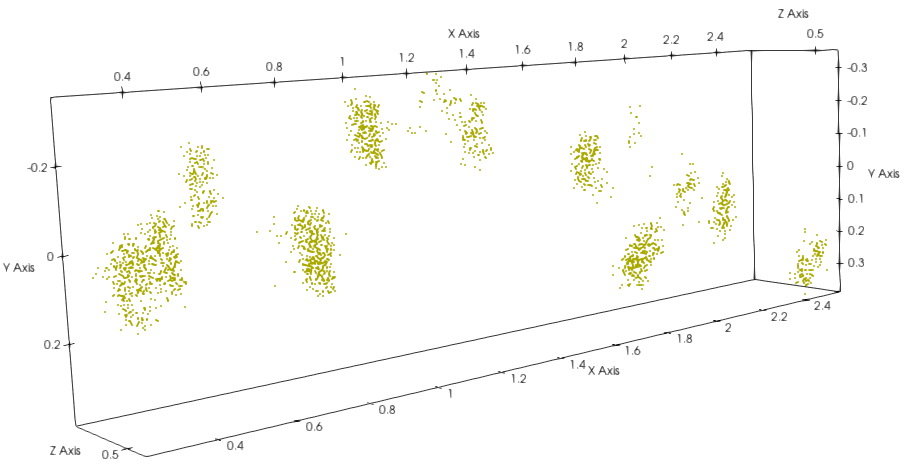}}
\caption{\textit{An example of a reconstructed 3D scene. The first 1 meter is reconstructed from the sequence of images in Fig. \ref{fig:series}.}}
\label{fig:example_3D}
\end{figure}

Due to the potential false positives and false negatives produced by Faster-RCCN as well as the filtering of matches described in section III.C.3, the final optimized pointcloud can contain outliers as well as missing seeds. After the retrieval of the optimized camera poses and landmarks, we project the centers of all 2D segmented ellipses $\{s_i\}$ in each left stereo image to 3D world coordinates using equation \ref{eq:proj} to get the set of 3D points $\{X_{i}\}$. (If a center $s_i$ was filtered during the stereo matching process, we assign it the same depth as its closest stereo-matched 2D neighbor). We add $X_i$ to the final pointcloud if no point $X'$ exists such that $|X'-X_i| < T$ (where $T$ is set based on the average size of a sorghum seed known a priori). Finally, for each point $X_i$ in the pointcloud, we calculate the variance of the distance between $X_i$ and its $N$ closest neighbors and reject $X_i$ if the variance is greater than a threshold. The variance threshold and $N$ are both determined experimentally. An example of a generated 3D reconstruction is shown in Fig. \ref{fig:example_3D}.
\subsection{Summary of System}
In this section, we give a summary of the proposed system (Fig. \ref{fig:pipeline}): For every new stereo frame (time $t$),  the Detection-and-Segmentation block detects and segments out all seeds in the right and left stereo images and fits an ellipse to each segmented area.
The centers of all ellipses are extracted to obtain two sets of semantic keypoints $C_t^L$ from the left image and  $C_t^R$ from the right image, which are then matched using the Object-Level-Feature-Association block to obtain a set of stereo matches. The stereo matches are unprojected to 3D to get a pointcloud $PC_{t}$. Each point in the pointcloud corresponds to an unprojected center of one sorghum seed. We obtain the initial estimate of the pose $P_{t}$ at time $t$  by solving for the relative transformation $\tilde{P}$ between $PC_{t}$ and $PC_{t-1}$ using Iterative Closest Point (ICP) and constraining the rotation matrix to being identity. 
We also perform temporal feature association between the segmented seeds from the left image at time $t$ ($C_t^L$), and the left image at the previous timestep $t-1$ ($C_{t-1}^L$). The 3D location estimate of already seen landmarks is refined, and new landmarks are added to the factor graph. We then run a batch optimizer to obtain the estimated 3D location of all seeds and the camera trajectory up to time $t$.
\begin{figure*}[h]
\begin{center}
   \includegraphics[width=1\linewidth, height=2in]{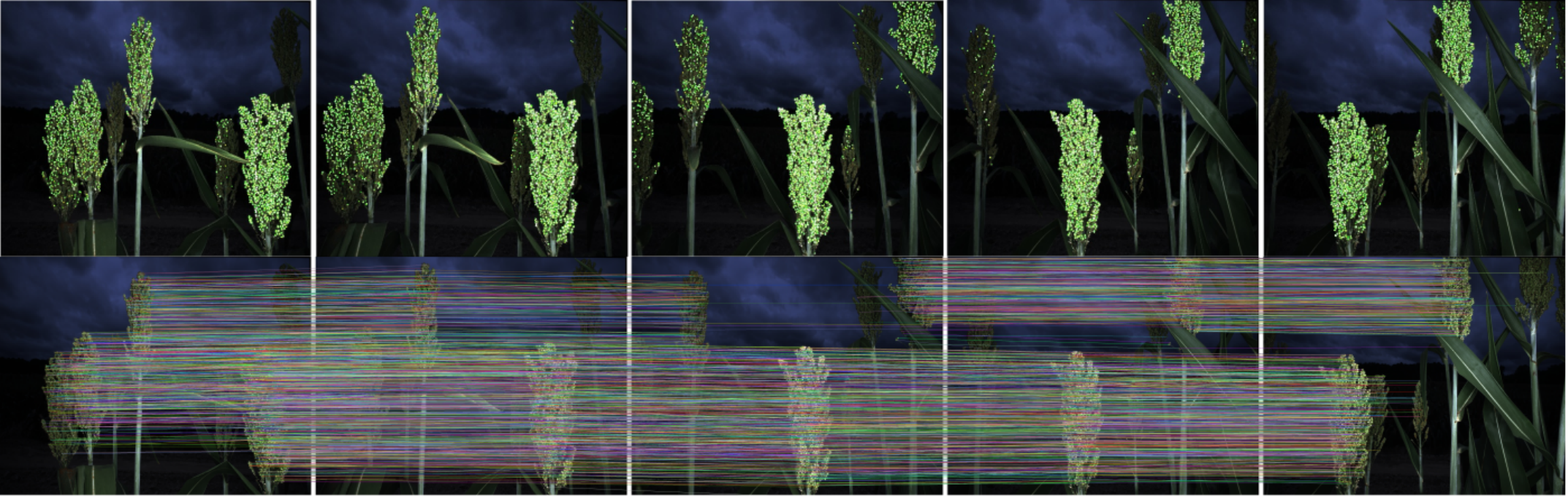}
\end{center}
   \caption{\textit{(a) The First row shows the bounding box detections. Each bounding box bounds one sorghum seed. The second row shows the output of the proposed data association pipeline for five consecutive images. (b) We note that the displacement of a projected 3D landmark between 2 consecutive frame is predominantly defined by a horizontal translation.}}
\label{fig:series}
\end{figure*}
\section{Results}
\begin{figure}[H]
\begin{center}
   \includegraphics[width=1.02\linewidth, height=4in]{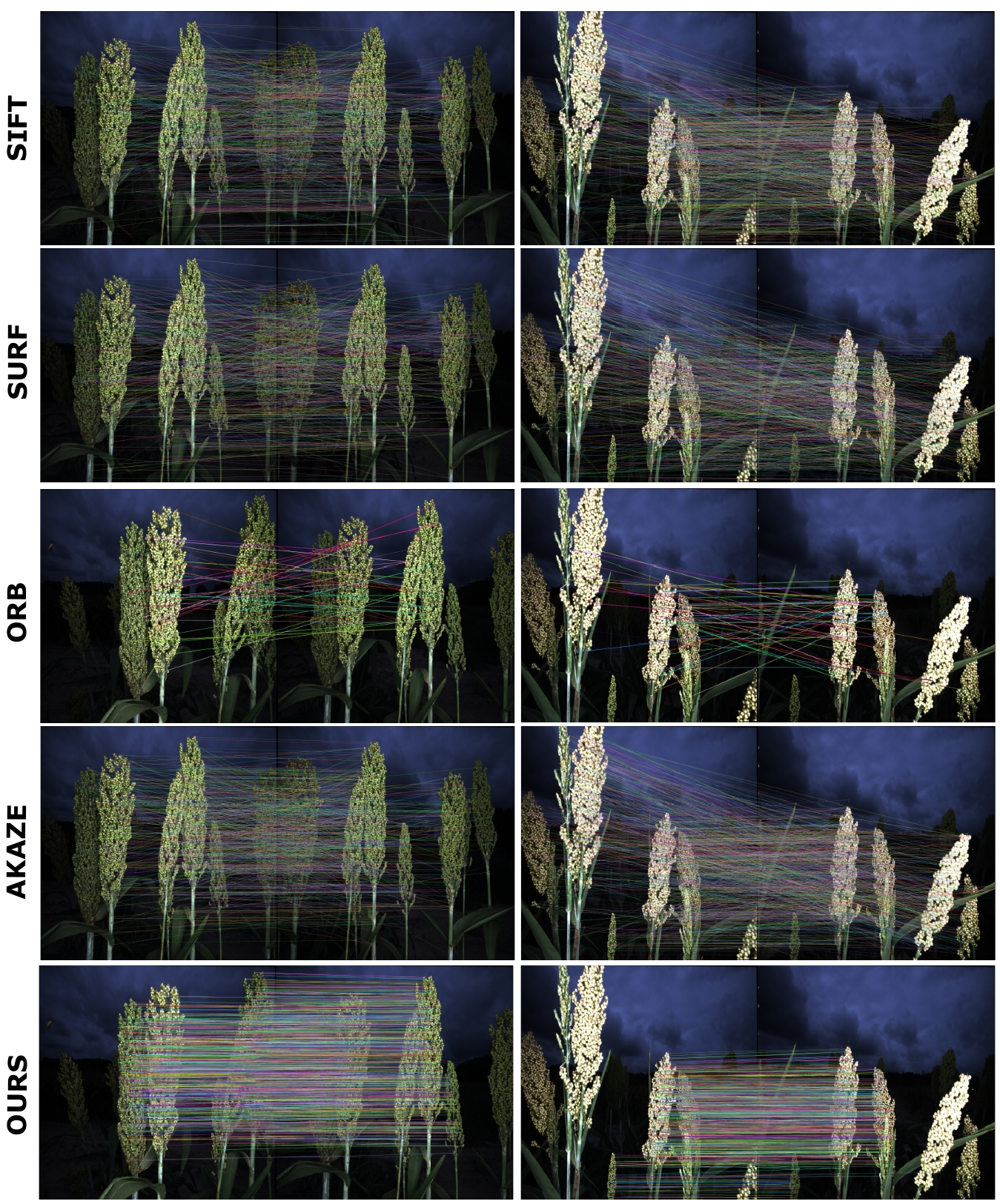}
\end{center}
   \caption{\textit{Examples of matches obtained using our data association algorithm versus matches obtained using the OpenCV implementations of SIFT, SURF, AKAZE, and ORB (before applying the motion constraints). Traditional feature detectors return a high number of incorrect matches.
   }}
\label{fig:example_output}
\end{figure}

We tested our system on a stereo image dataset of a sorghum field collected on August 21, 2018, in Florence, SC. It was captured using a custom stereo camera with a 0.11m baseline at a rate of 5Hz with an image resolution of $4096\text{px} \times 3000\text{px}$. The camera was mounted on a mobile field robot  \cite{robotanist}.  Sorghum fields are composed of rows, and each row is composed of several \textit{ranges}. A range is $\approx$ 4m long and may contain a different variety of sorghum. Empty spaces with no plants ($\approx 1.5m$ long) separate two consecutive ranges. 
\newline
First, we show in Fig. \ref{fig:example_output} examples of matches obtained using our data association algorithm versus matches obtained using the OpenCV implementations of the SIFT, SURF, AKAZE, and ORB algorithms with a Brute Force Descriptor Matcher. 
Fig. \ref{fig:example_output} shows that our data association algorithm produces accurate semantic matches, while the remaining four algorithms return a high number of incorrect correspondences.
\subsection{Maximum Distance Mapped}
We use the \textit{Maximum Distance Mapped} as a comparison and indicator metric for the stability of SLAM systems and performance of data association algorithms. The presence of many erroneous matches or few detected features affect the performance and stability of SLAM algorithms. For example, in ORB-SLAM2, a small number of detected/matched features could cause the system to enter the ``lost" state. Particularly in agricultural settings, traditional feature detectors have low accuracy due to the presence of repeated patterns and lack of texture. In our setting, the problem is further exacerbated by the lack of inertial sensory measurements and by the low frame rate of the stereo camera. Hence, we consider the maximum distance that a SLAM system was able to map before such failures occur as our comparison metric of interest. We report the results obtained from testing on image sequences of 8  sorghum ranges.
First, we compare our feature-level data association with existing traditional feature extraction algorithms. We replaced the Object-Level-Feature-Association block (Fig. \ref{fig:data_association}) with the existing OpenCV keypoint detection implementations of SIFT, SURF, ORB, and AKAZE. We used a Brute force matcher for feature matching and filtered out the matches using D.Lowe's ratio test with a ratio set to $0.8$. We then also apply our prior constraints on the motion of the camera (i.e. use the same assumptions about the camera motion made in our proposed data association algorithm). We disregarded all matched keypoint pairs $(x_1, y_1), (x_2, y_2)$ where $|y_2 - y_1| > \epsilon'$ given our prior knowledge that the displacement of a keypoint between 2 consecutive frames should mostly be defined by a horizontal translation (Fig. \ref{fig:series}). We noticed that resizing the images and decreasing the image resolution worsens the OpenCV keypoint detection and matching algorithms' performance. Thus, we conduct all tests on full resolution images.
\begin{table}[h!]
\caption{\textit{Maximum distance mapped (in meters). All methods compared under the same camera motion assumptions. Ground truth distances are extracted from GPS readings}}
\begin{center}
\resizebox{1\linewidth}{!}{%
\begin{tabular}{|c|c|c|c|c|c|}
\hline
\textbf{} & \multicolumn{5}{|c|}{\textbf{Feature detector + Matcher }} \\
\cline{2-6} 
\textbf{} & \textbf{\textit{SIFT + BF}}& \textbf{\textit{SURF + BF}}& \textbf{\textit{ORB + BF}} & \textbf{\textit{AKAZE + BF}} & \textbf{\textit{OURS}}\\
\hline
\makecell{ \textbf{Range 1} \\ (3.56 m)  } & \makecell{ 1.86 m }& \makecell{ 1.55 m } &  \makecell{ 0.2 m } &  \makecell{ 1.55 m } &  \makecell{ \textbf{3.56 m} }\\
\hline
\makecell{ \textbf{Range 2} \\ (5.00 m)  } & \makecell{ 0.25 m }& \makecell{ 0.25 m} &  Failed &  \makecell{0.19 m  } &  \makecell{ \textbf{5.00 m}  }\\
\hline
\makecell{ \textbf{Range 3} \\ (4.42 m) } & \makecell{ 0.5 m }& \makecell{ 0.38 m } & Failed &  \makecell{ 0.38 m  } &  \makecell{ \textbf{2.85 m}  }\\
\hline
\makecell{ \textbf{Range 4} \\ (4.1 m)  } & \makecell{ 1.47m }& \makecell{ 0.6 m  } &  Failed &  \makecell{ 1.14 m } &  \makecell{ \textbf{2.31 m} }\\
\hline
\makecell{ \textbf{Range 5} \\ (4.78 m) } & \makecell{ 0.57 m }& \makecell{ 0.74 m } &  failed &  \makecell{ 0.74 m } &  \makecell{ \textbf{2.31 m}  }\\
\hline
\makecell{ \textbf{Range 6} \\ 3.94 m  } & \makecell{ 1.4 m }& \makecell{ 0.19 m } &  \makecell{ 0.11 m  } &  \makecell{ 0.45 m } &  \makecell{ \textbf{3.2 m} } \\
\hline
\makecell{ \textbf{Range 7} \\ 5.03 m } & \makecell{ 3.15 m }& \makecell{ 0.26 m } &  Failed &  \makecell{ 0.46 m } &  \makecell{ \textbf{3.72 m}  }\\
\hline
\makecell{ \textbf{Range 8} \\ 4.43 m } & \makecell{ 4.04 m }& \makecell{ 0.94 m } &  Failed &  \makecell{ 0.94 m } &  \makecell{ \textbf{4.43 m} }\\
\hline
\end{tabular}
\label{tab1}}
\end{center}
\end{table}
\par
Table \ref{tab1} shows the maximum distance mapped in meters that was achieved with each method.  Using our proposed feature matching algorithm, we can map 3 out of 8 ranges completely and map $65\%$ of the remaining 5 ranges on average ($78\%$ on average across the 8 sorghum ranges). SIFT performed the best out of the remaining four algorithms with which we are able to map around $38\%$ of the 8 sorghum ranges on average. The higher performance of our proposed method is primarily due to relying on the geometric relation between neighbouring seeds instead of on visual features.
\par
Object-Level SLAM systems that rely on prior object models or use specific representations for the landmarks (such ellipsoid) have limited applicability in our environment \cite{rols}. Therefore, we compare the performance of our SLAM algorithm against ORB-SLAM2, a state of the art feature-based SLAM system using DBoW2 for feature matching.
\vspace{-3px}
\begin{table}[H]
\caption{\textit{Maximum distance mapped - ORB-SLAM2 vs OURS}}
\begin{center}
\resizebox{0.7\linewidth}{!}{%
\begin{tabular}{|c|c|c|}
\hline
\textbf{} & \textbf{\textit{ORB-SLAM2}}& \textbf{\textit{OURS}}\\
\hline
\makecell{ \textbf{Range 1} (3.56m) } &  0.35m  &  \textbf{3.56m} \\
\makecell{ \textbf{Range 2} (5.00m) } &  0.25m  &  \textbf{5m} \\
\makecell{ \textbf{Range 3} (4.42m) } &  0.18m  &  \textbf{2.85m}
\\
\makecell{ \textbf{Range 4} (4.10m) } &  0.25m  &  \textbf{2.31m}
\\
\makecell{ \textbf{Range 5} (4.78m) } &  0.31m  &  \textbf{2.31m}
\\
\makecell{ \textbf{Range 6} (3.94m) } &  0.12m  &  \textbf{3.2m}
\\
\makecell{ \textbf{Range 7} (5.03m) } &  0.33m  &  \textbf{3.72m}
\\
\makecell{ \textbf{Range 8} (4.43m) } &  0.26m  &  \textbf{4.43m} \\
\hline
\end{tabular}
\label{tab2}}
\end{center}
\end{table}
\vspace{-3px}
We report in Table \ref{tab2} the distance mapped by ORB-SLAM2 before it enters the ``lost" state after which, the system tries to re-localize at each new frame. We see that ORB-SLAM2 cannot map more than $0.35m$ since very few keypoints could be detected and matched robustly across frames. This is consistent with the performance of ORB features when used as the frontend of our proposed SLAM system and is caused by the same reasons explained above (repeated patterns, lack of texture, and low camera frame rate). 
\vspace{-6px}
\section{Conclusion}
In this paper, we presented a SLAM system that takes advantage of the structure of agricultural fields and a data association framework that utilizes the segmented sorghum seeds as landmarks in the environment. We show that our object-level data association algorithm enables our SLAM system to perform significantly better than traditional feature detection and matching algorithms in a setting with low texture, repetitive patterns, and images captured at a slow frame rate. Our system also performs better than existing state-of-the-art Visual-SLAM algorithms such as ORB-SLAM2 in terms of maximum mapped distance in challenging agricultural environments. 
The layout of farms for other types of fruits and seeds is similar to that of a sorghum field. Hence, our SLAM system can be easily adapted to construct robust 3D models of other plantations and obtain accurate phenotyping data or serve as sensory input to other downstream robotics tasks. Directions for future work include improving our SLAM system to reconstruct larger-scale 3D models for different types of crops. In addition, due to the high computational complexity of the Hungarian Algorithm, the proposed method is better suited for offline-SLAM or in settings with few objects to match. Thus, another direction for future work includes investigating learning-based feature matching algorithms and their online performance in agricultural environments.  Future work also includes reasoning about occlusions to enhance data association accuracy and the quality of the SLAM solution. 

\section{Acknowledgments}
This work was supported in part by the DOE APRAE TERRA 3P project and USDA NIFA CPS 2020-67021-31531. Thanks to the team at Clemson University Pee Dee Research center for their assistance with data collection. 
\nocite{*}
{
\bibliographystyle{IEEEtran}
\IEEEtriggeratref{36}
\bibliography{references}
}
\end{document}